%% file: main_arxiv.tex
\documentclass[letterpaper]{article} 
\usepackage{aaai2026}  
\usepackage{times}  
\usepackage{helvet}  
\usepackage{courier}  
\usepackage[hyphens]{url}  
\usepackage{graphicx} 
\usepackage{natbib}  
\usepackage{caption} 
\usepackage{algorithm}
\usepackage{algorithmic}

\usepackage{newfloat}
\usepackage{listings}
\DeclareCaptionStyle{ruled}{labelfont=normalfont,labelsep=colon,strut=off} 
\floatstyle{ruled}
\newfloat{listing}{tb}{lst}{}
\floatname{listing}{Listing}
\usepackage{xspace}
\usepackage{amsmath}
\usepackage{amssymb}
\usepackage[capitalize]{cleveref}
\usepackage{graphicx}
\usepackage{multirow}
\usepackage{colortbl}
\usepackage{booktabs}
\usepackage{xspace}
\usepackage{pifont}
\usepackage{caption}

\usepackage{lipsum} 

\crefname{section}{Sec.}{Secs.}
\Crefname{section}{Section}{Sections}
\Crefname{table}{Table}{Tables}
\crefname{table}{Tab.}{Tabs.}

\makeatletter
\DeclareRobustCommand\onedot{\futurelet\@let@token\@onedot}
\def\@onedot{\ifx\@let@token.\else.\null\fi\xspace}

\def\eg{\emph{e.g}\onedot} 
\def\ie{\emph{i.e}\onedot}

\def\wrt{w.r.t\onedot}

\makeatother

\def\DELIVER{\textbf{DELIVER}\xspace}
\def\MCubeS{\textbf{MCubeS}\xspace}
\def\MUSES{\textbf{MUSES}\xspace}
\def\wrtSOTA{\emph{\wrt SOTA}\xspace}

\def\CMNeXt{CMNeXt\xspace}
\def\MAGIC{MAGIC\xspace}
\def\AnySega{Any2Seg\xspace}
\def\AnySegb{AnySeg\xspace}

\title{CHARM: Collaborative Harmonization across Arbitrary Modalities for Modality-agnostic Semantic Segmentation}
\author {
    Lekang Wen\textsuperscript{\rm 1},
    Jing Xiao\textsuperscript{\rm 2}\thanks{Corresponding author},
    Liang Liao\textsuperscript{\rm 3},
    Jiajun Chen\textsuperscript{\rm 1},
    Mi Wang\textsuperscript{\rm 1},
}
\affiliations {
    \textsuperscript{\rm 1}State Key Laboratory of Information Engineering in Surveying, Mapping and Remote Sensing, Wuhan University\\
    \textsuperscript{\rm 2}School of Artificial Intelligence\\
    \textsuperscript{\rm 3}Hangzhou Institute of Technology, Xidian University\\
    wenlk3@whu.edu.com, jing@whu.edu.com, liaoliang1@xidian.edu.cn, jiajunchen@whu.edu.com, wangmi@whu.edu.com,
}

\begin{document}
\maketitle

\begin{abstract}
Modality-agnostic Semantic Segmentation (MaSS) aims to achieve robust scene understanding across arbitrary combinations of input modality. Existing methods typically rely on explicit feature alignment to achieve modal homogenization, which dilutes the distinctive strengths of each modality and destroys their inherent complementarity. To achieve cooperative harmonization rather than homogenization, we propose CHARM, a novel complementary learning framework designed to implicitly align content while preserving modality-specific advantages through two components: (1) Mutual Perception Unit (MPU), enabling implicit alignment through window-based cross-modal interaction, where modalities serve as both queries and contexts for each other to discover modality-interactive correspondences; (2) A dual-path optimization strategy that decouples training into Collaborative Learning Strategy (CoL) for complementary fusion learning and Individual Enhancement Strategy (InE) for protected modality-specific optimization. Experiments across multiple datasets and backbones indicate that CHARM consistently outperform the baselines, with significant increment on the fragile modalities. This work shifts the focus from model homogenization to harmonization, enabling cross-modal complementarity for true harmony in diversity.
\end{abstract}

\begin{figure*}[t]
    \centering
    \includegraphics[width=1\linewidth]{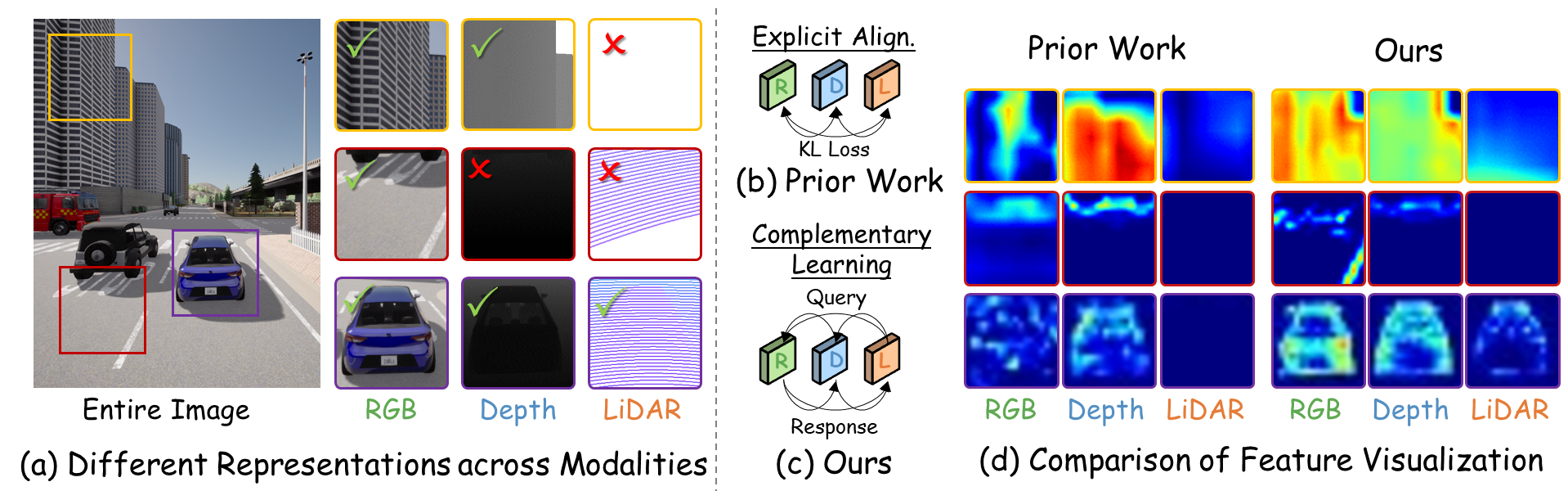}
    \vspace{-2mm}
    \caption{Example of modality misalignment and comparison of designed methods for MaSS. (a) Examples of content variations in different modalities; (b) Prior works use explicit alignment (\eg, KL divergence) that forces feature homogenization; (c) Our approach employs complementary learning through query and response to discover modality-interactive correspondences; (d) features of RGB and LiDAR are suppressed in prior work while in our work the features show their boosted specific information. }
    \label{fig: story}
    \vspace{-2mm}
\end{figure*}

\section{Introduction}

Multimodal Semantic Segmentation (MSS) aims to leverage complementary features from multiple visual modalities to achieve robust scene understanding, particularly under adverse conditions, \eg, heavy fog, nighttime, and rain. 
For instance, in autonomous driving scenarios under rainy conditions, RGB-based perception often suffers from substantial degradation due to reduced visibility. In contrast, auxiliary modalities such as LiDAR and Depth sensors can provide geometry-aware cues that compensate for the perception limitations of RGB, thereby improving driving safety. Most existing MSS methods follow an \textit{X}+\textit{A} paradigm, where \textit{X} denotes as a designated primary modality and \textit{A} represents one~\cite{tan2024epmf,dong2024egfnet, wei2025hdbformer,zhang2025cmffn,yang2025asymmetric} or multiple auxiliary modalities~\cite{zhang2023cmx,brodermann2023hrfuser,li2024stitchfusion,wan2024sigma,zhang2025tagfusion}. However, this paradigm inherently prioritizes the primary modality, which suppresses the contribution of auxiliary modalities, leading to significant performance degradation when the primary modality is unavailable or compromised.

Modality-agnostic Semantic Segmentation (MaSS) has emerged to enable robust performance under diverse and incomplete modality configurations. Early efforts tackled multimodal imbalance by employing modality dropout~\cite{zhang2022mmformer,zhang2023delivering,shi2024passion,maheshwari2024missing}, which randomly omits input sources during training. However, it may be insufficient to ensure effective representation learning for non-dominant or fragile modalities. 
Recent methods~\cite{chen2024novel,zheng2025learninga,zheng2024magic,zheng2025centering,zheng2025learning} have shifted toward feature alignment, enforcing cross-modal feature consistency through explicit constraints, \textit{e.g.}, KL divergence. MAGIC~\cite{zheng2024magic} groups modalities into robust and fragile sets and aligns their features with a cosine-based loss. Any2Seg~\cite{zheng2025learninga} incorporates semantic guidance from vision-language models to enhance alignment, while AnySeg~\cite{zheng2025learning} distills all modality features toward informative representation, improving performance in fragile-modal combinations. Although these alignment-based methods have shown to better handle combinations involving robust modalities, their reliance on explicit alignment often suppress modality-specific characteristics, leading to homogenized representations and reduced cross-modal complementarity.

As illustrated in \cref{fig: story}(a), different modalities could capture distinct aspects of the same scene, resulting in inherent misalignment. For example, the building missed in LiDAR while the Lane marking is only available in RGB. Explicit alignment-based methods as shown in (b) homogenizes these multimodal features by enforcing convergence across inherently divergent distributions. This process weakens modality-specific strengths, as RGB's rich information is diluted to match achromatic patterns of LiDAR for the building case, and LiDAR's precise spatial cues are suppressed in the car case as shown in (d). 
This limitation motivates us to pursue a novel approach that preserves the distinct characteristics of each modality while enabling effective collaboration, namely to harmonize modalities under two principles: 
1) modalities should coordinate through mutual understanding of each other, enabling the system to jointly model comprehensive scene information across modalities; 
2) the complementary features of each modality should be actively stimulated and reinforced during the multi-modal interaction, rather than suppressed by uniform alignment constraints. 




In this work, we propose \textbf{C}ollaborative \textbf{H}armonization across \textbf{AR}bitrary \textbf{M}odalities (CHARM), a MaSS framework that enables synergistic feature interaction while preserving modality-specific strengths.
The core of CHARM lies in two aspects according to the above principles. Firstly, a fundamental unit, named modality Mutual Perception Unit (MPU), is proposed to explore the complementarity
among modalities through cross-modal attention, where each modality simultaneously serve as queries and contexts to the others, enabling the discovery of modality-interactive correspondences at each feature scale. 
Second, two pathways are designed to systematically balance collaborative learning of all modalities (principle 1) with individual enhancement (principle 2): Collaboration Learning strategy (CoL) for joint optimization that dynamically adapts cooperation based on each modality's robustness, and Individual Enhancement strategy (InE) that provides protected learning spaces with the robustness guidance for individual modalities to stimulate their full potential. 





The main contributions of this work are threefold: 
\begin{itemize}
\item We identify the fundamental limitation of modality homogenization in existing explicit alignment strategy for MaSS on suppressing the modal-specific features, and propose a cooperative paradigm that enables effective complementarity across modalities; 


\item We designed a complementary learning framework, \textit{i.e.}, CHARM, with a fundamental unit MPU to explore the complementray across modalities, and two strategies of CoL for collaborative learning of all modalities and InE for individual enhancement by mutually learning from other modality;

\item Extensive experiments demonstrate that CHARM significantly improves the effectiveness of each modality and their various combinations, ensuring robust MaSS performance that consistently surpasses advanced baselines.

\end{itemize}

\section{Related Work}

\subsection{Multimodal Semantic Segmentation}

MSS aims to enhance segmentation accuracy and robustness by integrating information from different sensors. Previous approaches focused on fusing RGB with auxiliary modalities, such as Depth~\cite{feng2024ingredientguided,wei2025hdbformer}, Event~\cite{alonso2019evsegnet,xia2023cmda}, LiDAR~\cite{tan2024epmf,zhang2025cmffn}, thermal~\cite{dong2024egfnet,yang2025asymmetric}, and polarization~\cite{mei2022glass}. With the advancement of sensor technologies, dual-modal semantic segmentation has expanded to multimodal semantic segmentation, where approaches can be categorized into symmetric branch distribution~\cite{liang2022multimodal,reza2024mmsformer,li2024stitchfusion} and asymmetric branch distribution~\cite{brodermann2023hrfuser,wan2024sigma,zhang2025tagfusion} architectures. The former treats all modalities equally, such as  MMSFormer~\cite{reza2024mmsformer}, which achieves semantic segmentation by concatenating all multimodal features and fusing them through linear layers. The latter designates the most robust modality as the primary modality while treating others as auxiliary inputs. For instance, GFBN~\cite{gao2024global} uses RGB as the primary modality and incorporates other modalities to achieve arbitrary-modal semantic segmentation.

MSS methods demonstrate exceptional performance when all input modalities are in the normal state. However, they are generally designed for a specific modality combination, limiting their performance when applied to MaSS with any modalities missing or degraded.

\subsection{Modality-agnostic Semantic Segmentation}

MaSS aims to achieve robust performance across diverse and incomplete modality configurations. Initial approaches addressed multimodal imbalance through modality dropout strategies~\cite{sharma2020missing,zhang2022mmformer,zhang2023delivering,shi2024passion,maheshwari2024missing}, which randomly exclude input modalities during training. Nevertheless, such strategies may prove inadequate for ensuring effective representation learning of non-dominant or vulnerable modalities. Contemporary methods~\cite{wang2023learnable,liu2024fourier,chen2024novel,zheng2025learninga,zheng2024magic,zheng2025centering,zheng2025learning} have pivoted toward feature alignment paradigms, imposing cross-modal feature consistency via explicit constraints such as KL divergence. Notably,  MAGIC~\cite{zheng2024magic} categorizes modalities into robust and fragile subsets, aligning their features using cosine-based loss functions. Any2Seg~\cite{zheng2025learninga} leverages semantic guidance from vision-language models to enhance cross-modal alignment, whereas AnySeg~\cite{zheng2025learning} employs knowledge distillation to guide all modality features toward more informative representations, thereby improving performance in fragile-modal combinations. 

While current alignment-centric approaches demonstrate superior handling of combinations involving robust modalities, their dependence on explicit alignment tends to suppress modality-specific characteristics. This leads to homogenized representations that diminish cross-modal complementarity, potentially limiting the full exploitation of multimodal information.

\section{Methodology}

\begin{figure*}[tb]
    \centering
    \includegraphics[width=1\linewidth]{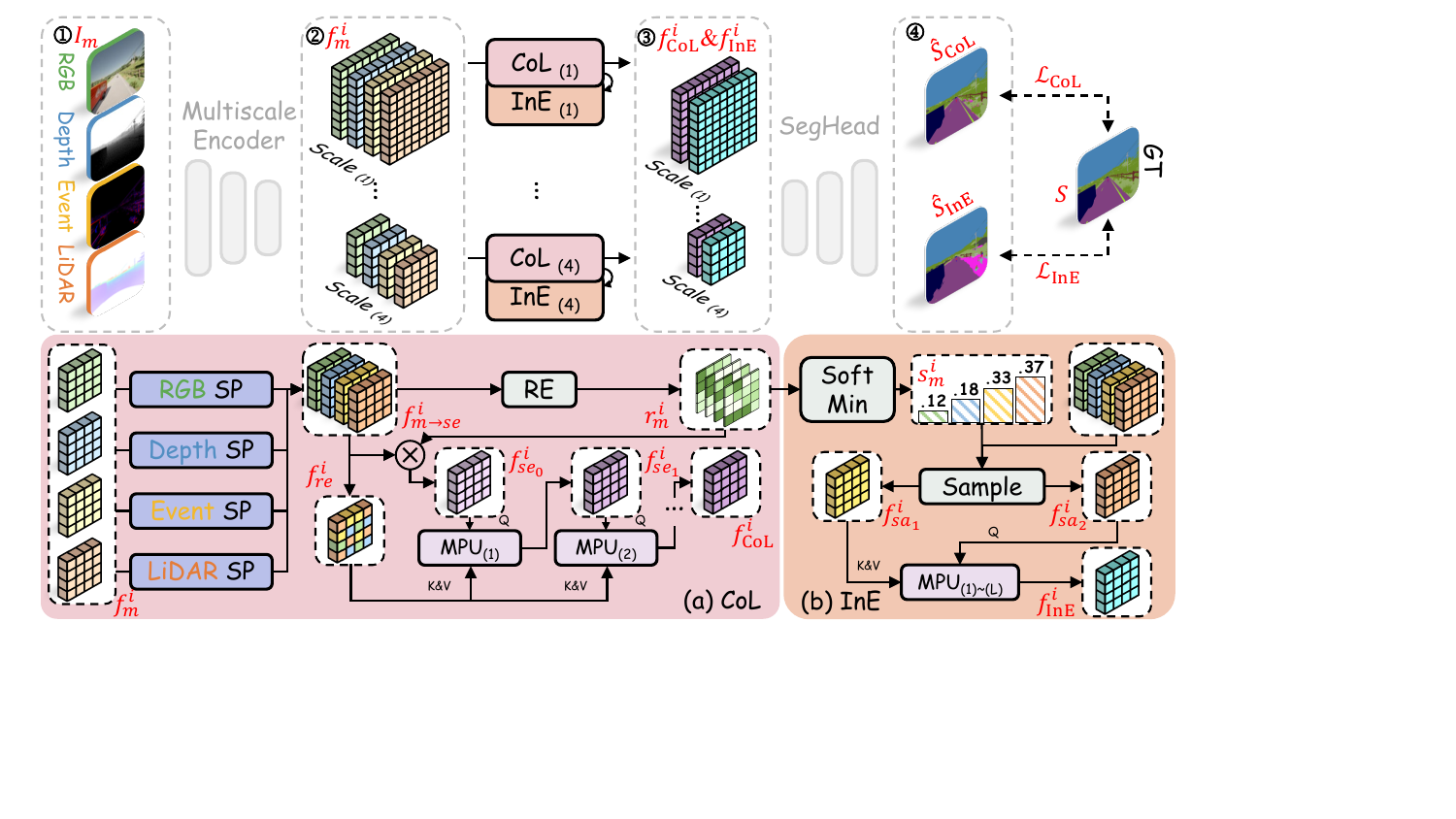}
    \caption{The overall framework of CHARM. At each scale, the encoder extracts multimodal features. CoL and InE are coupled when training and added between scales to fuse them progressively and input into the SegHead for segmentation. }
    \label{fig: framework}
    \vspace{-4mm}
\end{figure*}

\subsection{Problem Formulation of MaSS}


Following the common setting, we consider a modality set $\mathcal{M}$ comprising $M$ modalities. Let $\mathcal{I} = \{I_m\}_{m \in \mathcal{M}}$ represent the full multimodal input, and $S$ denote the corresponding segmentation labels. During training, all modalities are assumed to be accessible, while during inference, only a subset $\mathcal{I}^{\text{sub}} \subseteq \mathcal{I}$ is available due to sensor failures, environmental variability, or cost constraints. The goal of MaSS is to learn a segmentation model that maintains robust and accurate under arbitrary modality combinations, particularly in the presence of missing modalities.

Typically, models trained with full multimodal supervision suffers significant performance degradation when certain modalities are missing. To mitigate this issue, existing approaches employ a twofold solution: (1) random modality dropout during training to simulate diverse incomplete modality scenarios, generating input subsets $\mathcal{I}_{\text{sub}}$, and (2) cross-modal feature alignment to ensure representation consistency across modalities. The model $\Phi$ is then optimized to handle various modal combinations by minimizing:
\begin{equation}
    \begin{aligned}
        \min_{\Phi} \mathcal{L}_{\text{MaSS}} =\ 
        &\lambda_\text{seg} \mathcal{L}_{\text{seg}}(\Phi(\mathcal{I}_{\text{sub}}), S) \\
        +&\lambda_\text{align} \sum_{(m,m') \in \mathcal{P}} \mathcal{L}_{\text{align}}(f_m, f_{m'})),
    \end{aligned}
\end{equation}
where $\mathcal{P} = \{(m,m') | m,m' \in \mathcal{M}, m \neq m'\}$ represents all distinct modality pairs, $f_m$ and $f_{m'}$ denote the features from modalities $m$ and $m'$, $\mathcal{L}_{\text{seg}}$ is the cross-entropy loss for segmentation, $\mathcal{L}_{\text{align}}$ is the similarity loss for multimodal feature alignment, and $\lambda_\text{seg}$, $\lambda_\text{align}$ are balancing coefficients.

\subsection{Framework Overview}

The framework of the proposed CHARM is shown in \cref{fig: framework}, where we formulate MaSS as a cooperative harmonization problem and solve it with a two-pathway synergistic process. MPU is the core module that can accept arbitrary numbers and types of multimodal features as queries and contexts, enabling each querying modality to perceive the content of others during learning. It facilitates the discovery of modality-interactive correspondences, resulting in implicit alignment without explicit constraints. These aligned modalities can, in turn, complement and enhance each other.

To leverage MPU effectively, CHARM integrates two cooperative pathways: CoL uses extracted multimodal features as queries and contexts to mine modality-interactive correspondences between multiple modalities, enabling model to learn complementary information and maximize synergy across modalities; InE employs individual modal features as queries and contexts to mine interactive responses within single modalities. Although CoL effectively enhances complementary capabilities, the issue of modal imbalance potentially suppress the fragile modalities. In contrast, InE provides a protective learning space for all modalities, thereby improving the individual potential of each modality. 

Taking four modalities of RGB (R), Depth (D), Event (E), and LiDAR (L) modalities as an example, CHARM packs all input images $I_{m}$ into a mini-batch for efficient parallel computation. A shared-weight encoder $F$ extract features for each modality independently across each scale:  
\begin{equation}
    \{f_{m}^{i}\}_{i=1}^4=F(I_{m}).
\end{equation}
It is important to note that CoL and InE operate in couples: the feature $f_{m}^{i}$ from each modality is processed through both pathways to generate a modal-collaborative feature $f_{\text{CoL}}^i$ and a modal-enhanced feature $f_{\text{InE}}^i$. There features are then fed into a segmentation head to produce the modal-collaborative prediction $\hat{S}_{\text{CoL}}$ and the modal-enhanced prediction $\hat{S}_{\text{InE}}$ for joint optimization, respectively. 

\subsection{Mutual Perception Unit}

To discover modality-interactive correspondences, MPU enables mutual perception across modalities through iterative query, where each modality's features serve as both queries ($Q$) and contexts ($K$, $V$) for the others. Designed to be modality-agnostic, MPU can robustly handle arbitrary modality combinations during inference.  For multi-modal imagery, it introduce a windowed attention design to balance a large cross-modal receptive field with computational efficiency.

\begin{figure}
    \centering
    \includegraphics[width=1\linewidth]{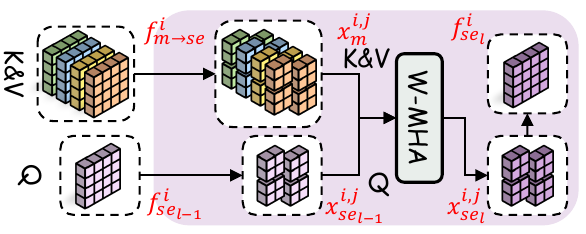}
    \caption{Structure of Mutual Perception Unit (MPU).}
    \label{fig: msa_block}
    \vspace{-2mm}
\end{figure}

To further enhance cross-modal interaction, multiple MPU blocks are employed in CHARM, where each block follows a specific structure shown in \cref{fig: msa_block}. Specifically, $f_{m\rightarrow se}^i$ is partitioned into $J$ windowed modal semantic features $\{x_{m}^{i,j}\}_{j=1}^J$, while $f_{se_{l-1}}^i$ is partitioned into $J$ windowed semantic features $\{x_{se}^{i,j}\}_{j=1}^J$. Each $\{x_{m}^{i,j}\}_{j=1}^J$ has modality-specific weight matrices for generating Key and Value, thus the generation process of $Q,K,V$ is formulated as:
\begin{equation}
\begin{aligned}
Q_{i,j} &= x_{se}^{i,j}W^Q,\\
K_{i,j} &= \{x_m^{i,j}W_{m}^K\}_{m\in\mathcal{M}},\\
V_{i,j} &= \{x_m^{i,j}W_{m}^V\}_{m\in\mathcal{M}}.  
\end{aligned}
\end{equation}
These serve as inputs for the Window Multi-Head Attention (W-MHA) \cite{liu2021swin}, outputting interacted windowed semantic features $\{x_{se}^{i,j}\}$, which are finally assembled through reverse partition to obtain the semantic feature $f_{se_{l}}^i$. Shifted window partitioning is introduced to enable cross-window connections by shifting during adjacent MPU blocks. Different from addition or concatenation fusion, MPU naturally accommodates arbitrary numbers of modalities as input, and the separable $Q$, $K$, $V$ relationships prevent any single modality from dominating the learning process, ensuring balanced multimodal interaction. In the both pathways described in the following, $L$ MPU blocks are embedded within both pathways at each scale. In CoL, MPU optimally integrates simultaneous multi-modal inputs, while in InE, it enables each modality to learn modality-interactive correspondences.

\subsection{Collaboration Learning Strategy}

\begin{figure}
    \centering
    \includegraphics[width=1\linewidth]{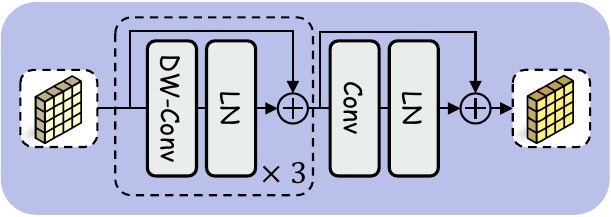}
    \caption{Structure of Semantic Projector (SP).}
    \label{fig: semantic_projector}
    \vspace{-2mm}
\end{figure}

As shown in \cref{fig: framework}(a), in CoL, the modal features $\{f_m^i\}_{m\in\mathcal{M}}$ at each scale are first projected to a semantic space via Semantic Projectors (SP), yielding $\{f_{m\rightarrow se}^i\}_{m\in\mathcal{M}}$. Each SP is a modality-specific sequential block consisting of three depth-wise convolutions with kernel sizes of $11 \times 11$, $7 \times 7$, and $3 \times 3$, followed by a $1 \times 1$ point-wise convolution, as described in \cref{fig: semantic_projector}. To quantify the reliabilit of each modality at every pixel, a Robustness Evaluator (RE), implemented as a $1 \times 1$ point-wise convolution, is used to compute the pixel-wise robustness scores $\{r_m^i\}_{m\in\mathcal{M}}$. These scores are then used to compute the initial semantic feature $f_{se_0}^i$ via a robustness-weighted summation across modalities:
\begin{equation}
    f_{se_0}^i = \sum_{m\in\mathcal{M}}{r_m^if_{m\rightarrow se}^i},
\end{equation}
which serves as the Query for the first MPU block. To reduce computational cost and modal dependency, modality-reassembled feature $f_{re}^i$ is randomly selected from the corresponding position across all modality-specific semantic features, serving as the Key and Value for MPU block to enable fine-grained, cross-model interaction. Through iterative processing across MPU blocks, the semantic features are progressively refined, ultimately generating the fused modal feature $f_\text{CoL}^i$. This design encourages complementary information exchange by enforcing cross-modal interactions, allowing modalities to compensate for each other's deficiencies in a content-aware and position-sensitive manner.

\subsection{Individual Enhancement Strategy}
As shown in \cref{fig: framework}(b), in InE, the modality robustness scores $\{r_m^i\}_{m\in\mathcal{M}}$ are averaged to obtain scalar values $\{\bar{r}_m^i\}_{m\in\mathcal{M}}$, which are then normalized through \textit{SoftMin} to derive sampling probabilities $\{s_m^i\}_{m\in\mathcal{M}}$:
\begin{equation}
    s_m^i = \textit{SoftMin}(\bar{r}_m^i), ~m\in\mathcal{M}.
\end{equation}

Subsequently, fragile modality-biased sampling is employed based on $\{s_m^i\}_{m\in\mathcal{M}}$ to extract complete modal semantic features $ f_{sa_1}^i$ and $ f_{sa_1}^i$ that serve as $Q,K,V$ for MPU blocks, generating the enhanced modal feature $f_\text{InE}^i$. This fragile modality-biased learning strategy emphasizes the potential of fragile modalities by providing them with more learning opportunities. Since the input modal features are complete and do not incorporate information from other modalities, InE implements protective learning for individual modalities. The combination of individual modal features and mutual perception facilitates modalities to discover complementary information from each other and promotes implicit alignment across modalities.

\subsection{Model Training and Inference}
As output features from CoL and InE have the same shape, they are packaged as input to SegHead for parallel segmentation. The loss functions $\mathcal{L}_\text{CoL}$ and $\mathcal{L}_\text{InE}$ both use cross-entropy loss for supervised training: 
\begin{equation}
    \begin{aligned}
        \mathcal{L}_\text{CoL}&=-\sum_{p\in S} S(p) \log{(\hat{S}_{\text{CoL}}(p))},\\
        \mathcal{L}_\text{InE}&=-\sum_{p\in S} S(p) \log{(\hat{S}_{\text{InE}}(p))},
    \end{aligned}
\end{equation}
where $p$ is the pixel index for segmentation map $S$. The MaSS optimization is then simplified as:
\begin{equation}
\min_{\Phi} \mathcal{L}_\text{MaSS} = \lambda_{\text{CoL}}\mathcal{L}_{\text{CoL}} + \lambda_{\text{InE}}\mathcal{L}_{\text{InE}},
\end{equation}
where $\mathcal{L}_{\text{CoL}}$ and $\mathcal{L}_{\text{InE}}$ denote the collaborative learning loss and modal enhancement loss, respectively. $\lambda_{\text{CoL}}$ and $\lambda_{\text{InE}}$ are balancing coefficients.

In inference phase, the model follows a process similar to CoL, with the difference that $f_{m\rightarrow se}^i$ is not transformed into $f_{re}^i$, but serves as the context for MPU, for full utilization of multimodal features. With features $\{f_{\text{CoL}}^i\}_{i=4}^4$ being input to SegHead, the predicted $\hat{S}_{\text{CoL}}$ serves as the final output.

\begin{figure*}[tb]
    \centering
    \includegraphics[width=\linewidth]{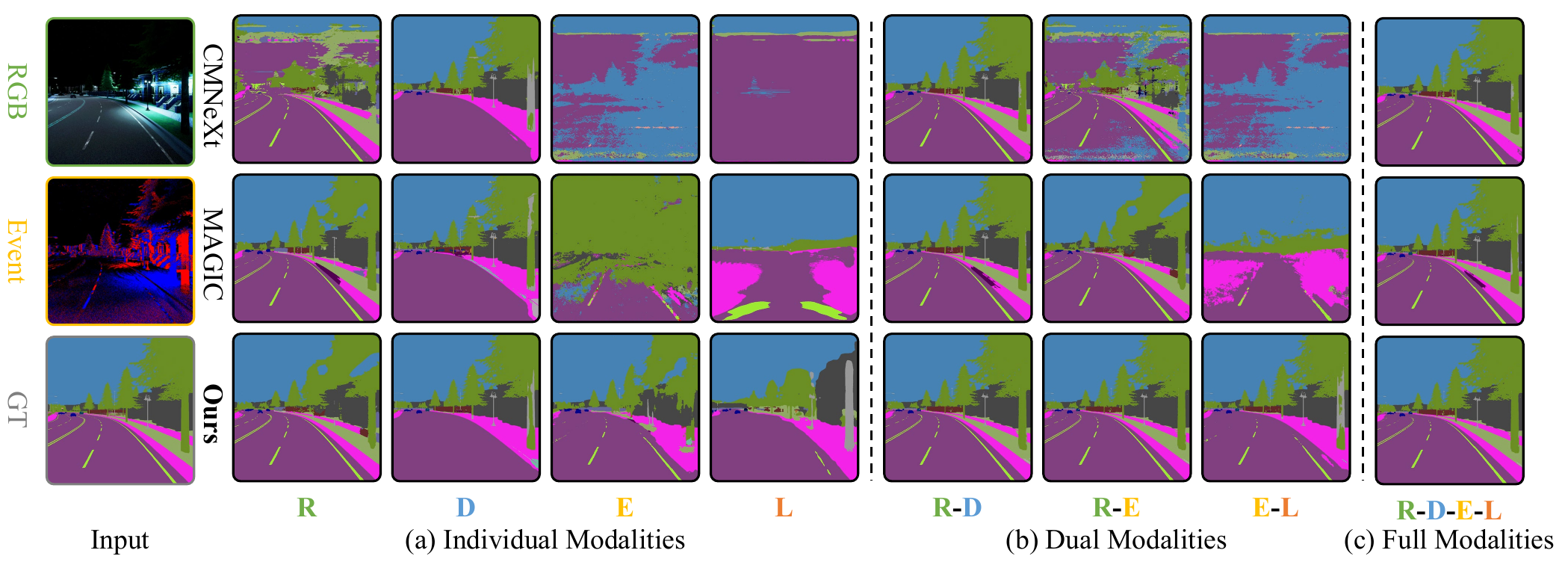}
    \caption{Visual quality comparisons of CMNeXt, MAGIC, and CHARM (Ours) on \DELIVER dataset of different situations and arbitrary-modal combinations.}
    \label{fig: deliver_result_visualize}
\end{figure*}

\begin{table*}[tb]
    \centering
    \caption{Objective quality comparison of methods trained with datasets: \DELIVER (four modalities), \MCubeS (four modalities), \MUSES (three modalities). \textbf{Bold} and \underline{underlined} letter the first and second best results. ``-'' denotes the absence of results due to the unavailability of \AnySega and \AnySegb. All the numbers in the table are the MIoU values (\%).}
    \label{tab: quantitative_comparisons}
    \resizebox{\linewidth}{!}{
    \input{tabs/quantitative_comparisons}}
\end{table*}

\begin{table*}[tb]
    \centering
    \caption{Ablation study on progressive component integration. The table demonstrates the cumulative benefits of each proposed module: (a) direct addition fusion baseline, (b) addition fusion with CoL, (c) addition fusion with CoL and InE, and (d) complete framework with all components. \textbf{Bold} and \underline{underlined} letter the first and second best results.}
    \label{tab: ablation}
    \resizebox{\linewidth}{!}{
    \input{tabs/ablation_studies}}
    \vspace{-4mm}
\end{table*}

\section{Experiments}
\subsection{Experimental Setup}

\noindent\textbf{Datasets.} We evaluate our method on three datasets: \DELIVER~\cite{zhang2023delivering} (general scenes with RGB, Depth, LiDAR, Event), \MCubeS~\cite{liang2022multimodal} (material segmentation with RGB, NIR, DoLP, AoLP), and \MUSES~\cite{brodermann2024muses} (driving scenarios with RGB, Event, LiDAR). 


\noindent\textbf{Baseline Methods.} We compare our method with four advanced methods, \ie, CMNeXt for MSS, and Any2Seg, MAGIC, AnySeg for MaSS. For equality, the backbones of all methods are set to MiT-B0 and MiT-B2 of Segformer~\cite{xie2021segformer}. We also conduct experiments with PVTv2~\cite{wang2022pvt} and Swin Transformer~\cite{liu2021swin} in the \textit{supplementary materials}. Note that results for AnySeg and Any2Seg are taken from their reports since there are no released codes or models.

\noindent\textbf{Metrics.} We evaluate performance using mean Intersection over Union (mIoU). \textit{Average}, \textit{Top-1}, and \textit{Last-1} denote the average, best, and worst mIoU across all-modal combinations, which assess overall performance, optimal complementary effects, and degree of fragile-modal potential activation, respectively. Detailed mIoU results for all combinations are provided in the \textit{supplementary materials}.

\subsection{Qualitative Comparison}

\cref{fig: deliver_result_visualize}(a)-(c) shows the qualitative comparisons of our method with the baseline methods on the \DELIVER dataset across different modality configurations. Results on other datasets are provided in \textit{supplementary materials}. 


\begin{itemize}
\item[(a)] \textbf{Individual Modalities}: Our method demonstrates consistent performance across all individual modalities by effectively exploiting the inherent potential of each input. It not only achieve better performance on robust modalities of RGB and Depth than the baselines, but also generate good results on fragile modalities of Event and LiDAR, on which the baselines completely fail to handle.  

\item[(b)] \textbf{Dual Modalities}: Our method exhibits superior cross-modal complementarity across various modality combinations. It achieves consistently high performance in robust+robust (R-D) settings. In robust+fragile (R-E) combinations, our method leverages Event information to compensate for RGB limitations, especially in dense tree foliage. Most notably, under the fragile+fragile setting (E-L), our method shows significant improvement in the tree region by leveraging complementary cues between Event and LiDAR.

\item[(c)] \textbf{Full Modalities}: When integrating all modalities (R-D-E-L), our method also achieves superior results. While baselines misclassify part of the tree trunk as utility poles, our method correctly identifies it as vegetation, demonstrating the significant complementary effects of our mutual perception mechanism.
\end{itemize}

\subsection{Quantitative Comparison}




\begin{figure}[tb]
    \centering
    \includegraphics[width=1\linewidth]{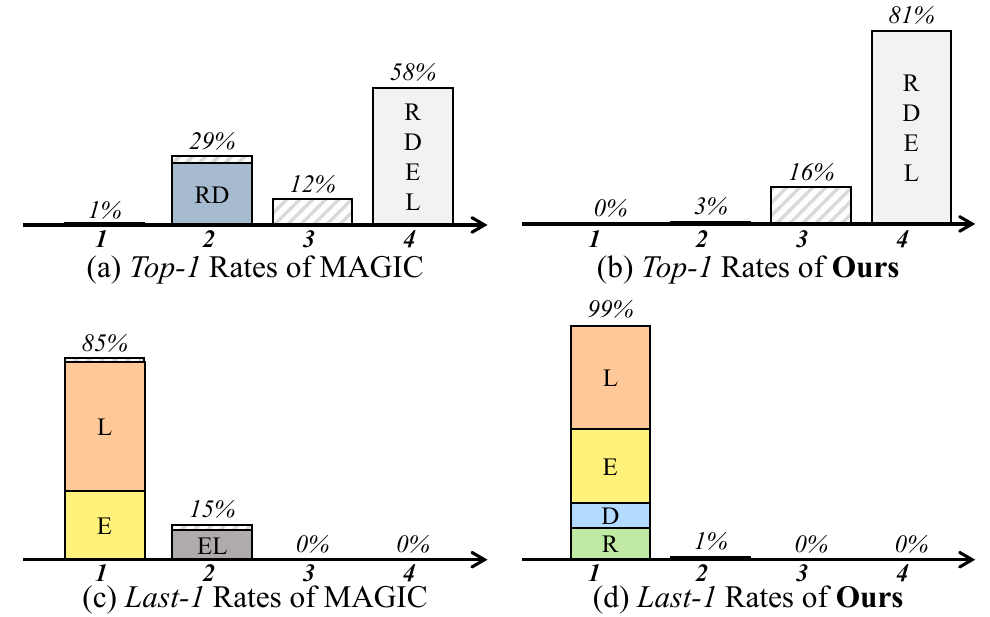}
    \caption{\textit{Top-1} (best performance) and \textit{Last-1} (worst performance) rates across modal combination sizes in \DELIVER.}
    \label{fig: argmax_visualize}
    \vspace{-4mm}
\end{figure}


We conduct quantitative comparisons on all datasets to evaluate the effectiveness of our method across arbitrary modality combinations, as show in \cref{tab: quantitative_comparisons}. It demonstrates that it achieves consistent improvements in \textit{Average}, \textit{Top-1}, and \textit{Last-1} mIoU over all baselines, validating its robustness in handling varying modality configurations. Specifically, on \DELIVER with MiT-B2, our method achieves 9.92\% gain in \textit{Average} mIoU over \AnySega. Notably, it also delivers notable improvements in handling fragile modalities, with \textit{Last-1} mIoU increasing by over 28.04\% across different backbones. These consistent improvements validate the effectiveness of our complementary learning approach in collaborative harmonization across modalities.

\cref{fig: argmax_visualize} further illustrate insights into modal combination performance by comparing our method with MAGIC. Firstly, the best results as shown in the \textit{Top-1} rates of all modality combinations are predominantly concentrated in combination of all four modalities, while the worst results shown by the \textit{Last-1} rates are more concentrate in the single modality in our method. This validates the effectiveness of the proposed CoL strategy in harnessing and integrating the complementary strengths of each modality, namely extra modality input can ensure better result.

Furthermore, the worst results was most in the fragile modalities of Event and Lidar from MAGIC. In our method, the worst results distribute more evenly in all modalities, owning to the reason that the InE strategy in our method has significantly boost the performance of fragile modalities by stimulating their own advantage and enhancing them through learning from other modalities.


\subsection{Ablation Studies}


\begin{figure}[tb]
    \centering
    \includegraphics[width=\linewidth]{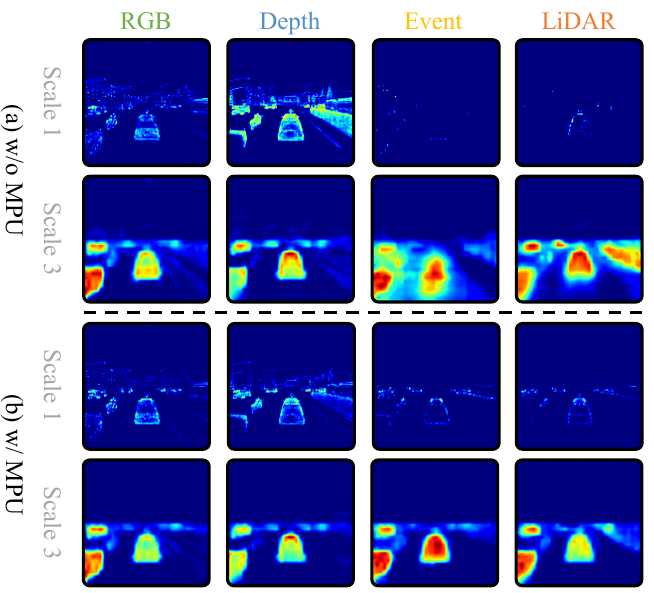}
    \caption{Feature visualization in \textbf{Variant (d)} without MPU and \textbf{Variant (e)} with MPU. }
    \label{fig: enhance_visualize}
    \vspace{-4mm}
\end{figure}


\noindent{\textbf{Progressive Component Analysis.}} The progressive integration of components reveals their distinct functional roles and indicates the superiority of MPU-based methods over simple additive strategies. \textbf{Variant (a)} serves as the baseline with direct additive fusion, suffering from severe modality imbalance as evidenced by poor \textit{Last-1} mIoU across all datasets. \textbf{Variant (b)} introduces CoL without MPU, showing modest improvements in average performance but still struggling with fragile-modal combinations. 
Comparing \textbf{Variant (c)} with \textbf{Variant (b)}, MPU-equipped methods achieve significant gains in \textit{Average} and \textit{Top-1} mIoU, demonstrating MPU's effectiveness in handling fragile-modal scenarios. This superiority stems from MPU's mutual perception mechanism that iteratively inherits complementary content from all modalities, preventing information-rich robust modalities from being diluted by information-sparse fragile modalities as occurs in additive fusion. \textbf{Variant (d)} incorporates InE alongside CoL, providing protective mechanism that further alleviate performance suppression of fragile modalities, validated by the improvements in \textit{Last-1} mIoU. \textbf{Variant (e)} achieves optimal performance by integrating all components, where MPU harmonizes cross-modal interactions while CoL enables robust complementary fusion and InE ensures individual modal enhancement. 

To further validate MPU's capability in collaborative harmonization, we visualize the features by Grad-CAM~\cite{selvaraju2017gradcam} in \textbf{Variant (d)} without MPU and \textbf{Variant (e)} with MPU in \cref{fig: enhance_visualize}. It shows that at Scale 1, robust modalities (RGB, Depth) show significantly improved on target vehicles with enhanced textural cues extraction, while fragile modalities (Event, LiDAR) show reduced noise and more concentrated responses. At Scale 3, all modalities exhibit substantially refined attention patterns, with Event and LiDAR showing markedly reduced diffusion in irrelevant regions while maintaining their unique semantic contributions. This cross-modal mutual perception mechanism enables each modality to leverage contextual information from others, leading to enhanced discriminative power for robust modalities and focused representational domains for fragile modalities. This validates that MPU enables maximum texture extraction in shallow layers and facilitates semantic exchange without mutual suppression in deeper layers, successfully achieving harmonization rather than homogenization while preserving modality-specific strengths.

\begin{figure}[tb]
    \centering
    \includegraphics[width=1\linewidth]{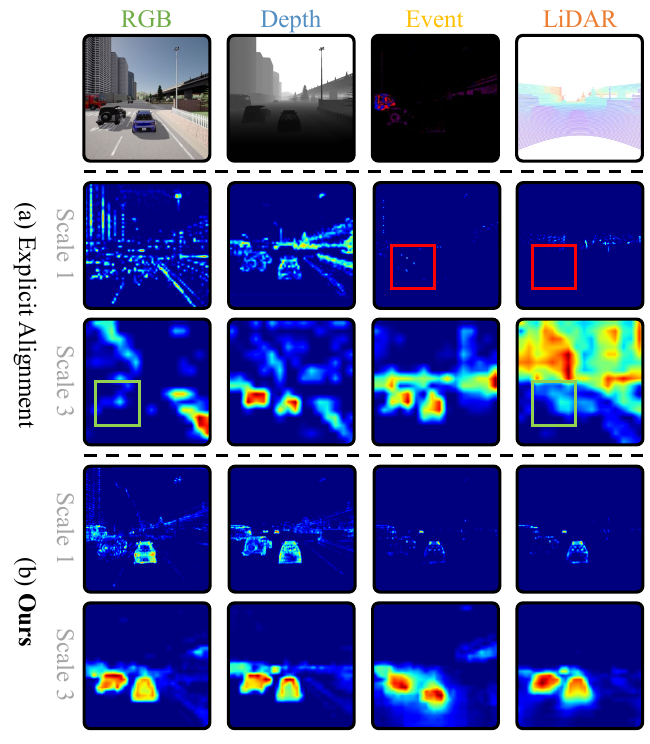}
    \caption{Feature visualization across different strategies. Colored boxes highlight different failure modes.}
    \label{fig: cam_visualize}
    \vspace{-4mm}
\end{figure}



\noindent\textbf{Comparison between MPU with Explicit Alignment.} To demonstrate the effectiveness of MPU in promoting implicit alignment, we replace the MPU by a explicit alignment that applies KL divergence loss at each scale to force multimodal features toward their average representation.
The visualization in \cref{fig: cam_visualize} reveals clear differences in activation patterns: explicit alignment unit forces all modalities toward centralized representations causing universal degradation. In contrast, MPU maintains distinctive yet complementary activation patterns across modalities. The red boxes illustrate ineffective learning in fragile modalities under explicit constraints, where the model captures features beyond its inherent capacity at Scale 1. The green boxes show degraded activations in robust modalities at Scale 3, with scattered and weakened responses due to forced alignment. In contrast, MPU preserves the representational integrity of each modality across both shallow and deep layers, enabling effective cross-modal cooperation without suppressing modality-specific strengths.

\begin{figure}[tb]
    \centering
    \includegraphics[width=\linewidth]{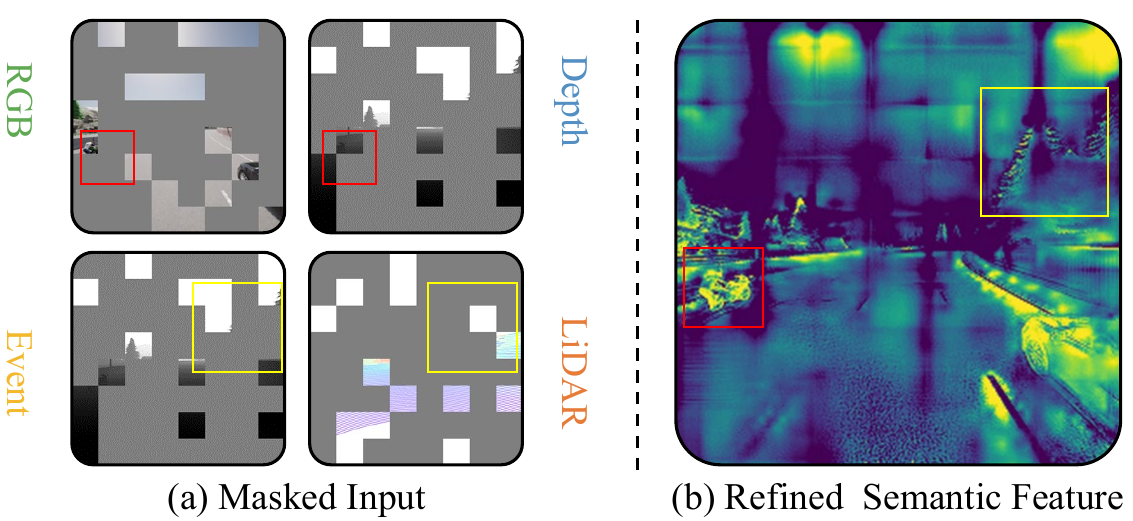}
    \caption{Visualization for multimodal complementarity. (a) Random masking inputs four modalities. (b) Feature visualization at Scale 3.}
    \label{fig: complementray_visualize}
    \vspace{-4mm}
\end{figure}

\noindent\textbf{Abilities in Complementary Fusion.} Our method enables complementary fusion by selectively integrating distinctive information from different modalities based on their inherent strengths. \cref{fig: complementray_visualize}(a) shows how our method effectively extracts and fuses key information using a masking strategy that retains only complementary information blocks from each modality. \cref{fig: complementray_visualize}(b) further illustrates the complementary fusion process: features of the motorcycle (red box) are primarily derived from RGB and Depth, leveraging their strong object recognition and spatial understanding capabilities, while features of the tree (yellow box) are captured from Event and LiDAR, which capture their temporal dynamics and precise geometry. This validates that our method enables maximum texture extraction in shallow layers and facilitates semantic exchange in deeper layers without suppressing modality-specific representations, achieving harmonization rather than homogenization.



\section{Conclusion}

This paper identify that current MaSS methods pursue multimodal homogenization, leading to diluted representations and suppressed complementary characteristics. To address it, we propose CHARM, a cooperative framework that achieves harmonization through two key innovations. The first is the fundamental MPU that enables discovering modality-interactive correspondences without explicit alignment constraints. Secondly, CoL and InE strategies provide systematic balance between collaborative learning and individual modal enhancement, ensuring optimal utilization of each modality's unique potential through robustness-guided cooperation. The consistent gains across different datasets and backbones confirm that our cooperative paradigm successfully preserves modal distinctiveness, and ensures the robust Mass performance by improving the effectiveness of each modality and various modality combinations in the modality-agnostic settings.

\bibliography{aaai2026}


\end{document}

%% file: tabs/quantitative_comparisons.tex
\begin{tabular}{c|l|ccc|ccc|ccc}
\hline
 & \multicolumn{1}{c|}{} & \multicolumn{3}{c|}{\DELIVER} & \multicolumn{3}{c|}{\MCubeS} & \multicolumn{3}{c}{\MUSES} \\ \cline{3-11} 
\multirow{-2}{*}{Backbones} & \multicolumn{1}{c|}{\multirow{-2}{*}{Methods}} & \textit{Average$\uparrow$} & \textit{Top-1$\uparrow$} & \textit{Last-1$\uparrow$} & \textit{Average$\uparrow$} & \textit{Top-1$\uparrow$} & \textit{Last-1$\uparrow$} & \textit{Average$\uparrow$} & \textit{Top-1$\uparrow$} & \textit{Last-1$\uparrow$} \\ \hline
 & \CMNeXt & 22.05 & 59.18 & 0.37 & 14.35 & 40.94 & 0.46 & 10.80 & 46.66 & 2.64 \\
 & \MAGIC & 40.49 & \underline{63.40} & 0.26 & \underline{34.56} & \underline{47.85} & \underline{0.55} & 33.34 & 49.05 & 2.68 \\
 & \AnySega & - & - & - & - & - & - & 33.86 & 50.00 & 3.17 \\
 & \AnySegb & \underline{47.51} & 59.72 & \underline{21.74} & - & - & - & \underline{40.23} & \underline{51.25} & \underline{19.57} \\
 & CHARM   (\textbf{Ours}) & \textbf{50.97} & \textbf{64.03} & \textbf{26.47} & \textbf{39.05} & \textbf{48.24} & \textbf{28.37} & \textbf{42.19} & \textbf{52.31} & \textbf{22.31} \\
\multirow{-6}{*}{MiT-B0} & {\color[HTML]{FF0000} \wrtSOTA} & {\color[HTML]{FF0000} \textit{+3.45}} & {\color[HTML]{FF0000} \textit{+0.63}} & {\color[HTML]{FF0000} \textit{+4.73}} & {\color[HTML]{FF0000} \textit{+4.49}} & {\color[HTML]{FF0000} \textit{+0.39}} & {\color[HTML]{FF0000} \textit{+27.82}} & {\color[HTML]{FF0000} \textit{+1.96}} & {\color[HTML]{FF0000} \textit{+1.06}} & {\color[HTML]{FF0000} \textit{+2.74}} \\ \hline
 & \CMNeXt & 25.15 & 66.43 & \underline{0.72} & 25.17 & 51.54 & \underline{1.54} & 33.14 & \underline{58.28} & 1.23 \\
 & \MAGIC & 44.66 & 67.66 & 0.27 & \underline{38.00} & \underline{53.01} & 0.32 & \underline{36.19} & 55.36 & \underline{3.34} \\
 & \AnySega & \underline{45.04} & \underline{68.25} & 0.31 & - & - & - & - & - & - \\
 & CHARM   (\textbf{Ours}) & \textbf{54.96} & \textbf{68.43} & \textbf{28.76} & \textbf{46.58} & \textbf{54.33} & \textbf{38.50} & \textbf{46.18} & \textbf{59.36} & \textbf{25.28} \\
\multirow{-5}{*}{Mit-B2} & {\color[HTML]{FF0000} \wrtSOTA} & {\color[HTML]{FF0000} \textit{+9.92}} & {\color[HTML]{FF0000} \textit{+0.18}} & {\color[HTML]{FF0000} \textit{+28.04}} & {\color[HTML]{FF0000} \textit{+8.57}} & {\color[HTML]{FF0000} \textit{+1.32}} & {\color[HTML]{FF0000} \textit{+36.96}} & {\color[HTML]{FF0000} \textit{+9.98}} & {\color[HTML]{FF0000} \textit{+1.08}} & {\color[HTML]{FF0000} \textit{+21.93}} \\ \hline
\end{tabular}%

%% file: tabs/ablation_studies.tex
\begin{tabular}{c|ccc|ccc|ccc|ccc}
\hline
\multirow{2}{*}{Variants} & \multicolumn{3}{c|}{Components} & \multicolumn{3}{c|}{DELIVER} & \multicolumn{3}{c|}{MCubeS} & \multicolumn{3}{c}{MUSES} \\ \cline{2-13} 
 & MPU & CoL & InE & \textit{Average$\uparrow$} & \textit{Top-1$\uparrow$} & \textit{Last-1$\uparrow$} & \textit{Average$\uparrow$} & \textit{Top-1$\uparrow$} & \textit{Last-1$\uparrow$} & \textit{Average$\uparrow$} & \textit{Top-1$\uparrow$} & \textit{Last-1$\uparrow$} \\ \hline
(a) & \ding{55} & \ding{55} & \ding{55} & 33.95 & 63.17 & 2.30 & 25.59 & 45.38 & 3.39 & 30.86 & 51.65 & 2.26 \\
(b) & \ding{55} & \ding{51} & \ding{55} & 41.53 & 62.50 & 3.28 & 33.38 & 46.83 & 14.21 & 31.38 & 51.39 & 0.81 \\
(c) & \ding{51} & \ding{51} & \ding{55} & 45.23 & \underline{63.97} & 15.89 & 35.22 & \underline{47.31} & 20.15 & 35.88 & \underline{51.96} & 15.20 \\
(d) & \ding{55} & \ding{51} & \ding{51} & \underline{48.76} & 62.78 & \underline{24.40} & \underline{36.70} & 46.92 & \underline{26.38} & \underline{38.67} & 51.77 & \underline{19.97} \\
(e) & \ding{51} & \ding{51} & \ding{51} & \textbf{50.97} & \textbf{64.03} & \textbf{26.47} & \textbf{39.05} & \textbf{48.24} & \textbf{28.37} & \textbf{42.19} & \textbf{52.31} & \textbf{22.31} \\ \hline
\end{tabular}%

%% file: main_arxiv.bbl
\begin{thebibliography}{34}
\providecommand{\natexlab}[1]{#1}

\bibitem[{Alonso and Murillo(2019)}]{alonso2019evsegnet}
Alonso, I.; and Murillo, A.~C. 2019.
\newblock {{EV-SegNet}}: {{Semantic Segmentation}} for {{Event-Based Cameras}}.
\newblock In \emph{2019 {{IEEE}}/{{CVF Conference}} on {{Computer Vision}} and
  {{Pattern Recognition Workshops}} ({{CVPRW}})}, 1624--1633.

\bibitem[{Br{\"o}dermann et~al.(2024)Br{\"o}dermann, Bruggemann, Sakaridis, Ta,
  Liagouris, Corkill, and Van~Gool}]{brodermann2024muses}
Br{\"o}dermann, T.; Bruggemann, D.; Sakaridis, C.; Ta, K.; Liagouris, O.;
  Corkill, J.; and Van~Gool, L. 2024.
\newblock {{MUSES}}: {{The Multi-sensor Semantic Perception Dataset}} for
  {{Driving Under Uncertainty}}.
\newblock In \emph{2024 European Conference on Computer Vision (ECCV)}, 21--38.

\bibitem[{Br{\"o}dermann et~al.(2023)Br{\"o}dermann, Sakaridis, Dai, and
  Van~Gool}]{brodermann2023hrfuser}
Br{\"o}dermann, T.; Sakaridis, C.; Dai, D.; and Van~Gool, L. 2023.
\newblock {{HRFuser}}: A Multi-Resolution Sensor Fusion Architecture for {{2D}}
  Object Detection.
\newblock In \emph{2023 {{IEEE}} 26th International Conference on Intelligent
  Transportation Systems ({{ITSC}})}, 4159--4166.

\bibitem[{Chen, Zhao, and Bruzzone(2024)}]{chen2024novel}
Chen, Y.; Zhao, M.; and Bruzzone, L. 2024.
\newblock A {{Novel Approach}} to {{Incomplete Multimodal Learning}} for
  {{Remote Sensing Data Fusion}}.
\newblock \emph{IEEE Transactions on Geoscience and Remote Sensing}, 62: 1--14.

\bibitem[{Dong et~al.(2024)Dong, Zhou, Xu, and Yan}]{dong2024egfnet}
Dong, S.; Zhou, W.; Xu, C.; and Yan, W. 2024.
\newblock {{EGFNet}}: {{Edge-aware}} Guidance Fusion Network for
  {{RGB}}--Thermal Urban Scene Parsing.
\newblock \emph{IEEE Transactions on Intelligent Transportation Systems},
  25(1): 657--669.

\bibitem[{Feng et~al.(2024)Feng, Xiong, Min, Hou, Duan, Liu, and
  Jiang}]{feng2024ingredientguided}
Feng, Z.; Xiong, H.; Min, W.; Hou, S.; Duan, H.; Liu, Z.; and Jiang, S. 2024.
\newblock Ingredient-Guided {{RGB-d}} Fusion Network for Nutritional
  Assessment.
\newblock \emph{IEEE Transactions on AgriFood Electronics}, 1--11.

\bibitem[{Gao et~al.(2024)Gao, Yang, Jiang, Fu, and Du}]{gao2024global}
Gao, S.; Yang, X.; Jiang, L.; Fu, Z.; and Du, J. 2024.
\newblock Global Feature-Based Multimodal Semantic Segmentation.
\newblock \emph{Pattern Recognition}, 151: 110340.

\bibitem[{Li et~al.(2024)Li, Zhang, Zhao, Gao, and Li}]{li2024stitchfusion}
Li, B.; Zhang, D.; Zhao, Z.; Gao, J.; and Li, X. 2024.
\newblock {{StitchFusion}}: {{Weaving Any Visual Modalities}} to {{Enhance
  Multimodal Semantic Segmentation}}.
\newblock arXiv:2408.01343.

\bibitem[{Liang et~al.(2022)Liang, Wakaki, Nobuhara, and
  Nishino}]{liang2022multimodal}
Liang, Y.; Wakaki, R.; Nobuhara, S.; and Nishino, K. 2022.
\newblock Multimodal {{Material Segmentation}}.
\newblock In \emph{2022 {{IEEE}}/{{CVF Conference}} on {{Computer Vision}} and
  {{Pattern Recognition}} ({{CVPR}})}, 19768--19776.

\bibitem[{Liu et~al.(2024)Liu, Zhang, Peng, Chen, Cao, Zheng, Sarfraz, Yang,
  and Stiefelhagen}]{liu2024fourier}
Liu, R.; Zhang, J.; Peng, K.; Chen, Y.; Cao, K.; Zheng, J.; Sarfraz, M.~S.;
  Yang, K.; and Stiefelhagen, R. 2024.
\newblock Fourier {{Prompt Tuning}} for {{Modality-Incomplete Scene
  Segmentation}}.
\newblock In \emph{2024 {{IEEE Intelligent Vehicles Symposium}} ({{IV}})},
  961--968.

\bibitem[{Liu et~al.(2021)Liu, Lin, Cao, Hu, Wei, Zhang, Lin, and
  Guo}]{liu2021swin}
Liu, Z.; Lin, Y.; Cao, Y.; Hu, H.; Wei, Y.; Zhang, Z.; Lin, S.; and Guo, B.
  2021.
\newblock Swin {{Transformer}}: {{Hierarchical Vision Transformer}} Using
  {{Shifted Windows}}.
\newblock In \emph{2021 {{IEEE}}/{{CVF International Conference}} on {{Computer
  Vision}} ({{ICCV}})}, 9992--10002.

\bibitem[{Maheshwari, Liu, and Kira(2024)}]{maheshwari2024missing}
Maheshwari, H.; Liu, Y.-C.; and Kira, Z. 2024.
\newblock Missing {{Modality Robustness}} in {{Semi-Supervised Multi-Modal
  Semantic Segmentation}}.
\newblock In \emph{2024 {{IEEE}}/{{CVF Winter Conference}} on {{Applications}}
  of {{Computer Vision}} ({{WACV}})}, 1009--1019.

\bibitem[{Mei et~al.(2022)Mei, Dong, Dong, Yang, Baek, Heide, Peers, Wei, and
  Yang}]{mei2022glass}
Mei, H.; Dong, B.; Dong, W.; Yang, J.; Baek, S.-H.; Heide, F.; Peers, P.; Wei,
  X.; and Yang, X. 2022.
\newblock Glass {{Segmentation}} Using {{Intensity}} and {{Spectral
  Polarization Cues}}.
\newblock In \emph{2022 {{IEEE}}/{{CVF Conference}} on {{Computer Vision}} and
  {{Pattern Recognition}} ({{CVPR}})}, 12612--12621.

\bibitem[{Reza, {Prater-Bennette}, and Asif(2024)}]{reza2024mmsformer}
Reza, M.~K.; {Prater-Bennette}, A.; and Asif, M.~S. 2024.
\newblock {{MMSFormer}}: {{Multimodal}} Transformer for Material and Semantic
  Segmentation.
\newblock \emph{IEEE Open Journal of Signal Processing}, 5: 599--610.

\bibitem[{Selvaraju et~al.(2017)Selvaraju, Cogswell, Das, Vedantam, Parikh, and
  Batra}]{selvaraju2017gradcam}
Selvaraju, R.~R.; Cogswell, M.; Das, A.; Vedantam, R.; Parikh, D.; and Batra,
  D. 2017.
\newblock Grad-{{CAM}}: {{Visual Explanations}} from {{Deep Networks}} via
  {{Gradient-Based Localization}}.
\newblock In \emph{2017 {{IEEE International Conference}} on {{Computer
  Vision}} ({{ICCV}})}, 618--626.

\bibitem[{Sharma and Hamarneh(2020)}]{sharma2020missing}
Sharma, A.; and Hamarneh, G. 2020.
\newblock Missing {{MRI Pulse Sequence Synthesis Using Multi-Modal Generative
  Adversarial Network}}.
\newblock \emph{IEEE Transactions on Medical Imaging}, 39(4): 1170--1183.

\bibitem[{Shi et~al.(2024)Shi, Shang, Sun, Yu, Yang, and Yan}]{shi2024passion}
Shi, J.; Shang, C.; Sun, Z.; Yu, L.; Yang, X.; and Yan, Z. 2024.
\newblock {{PASSION}}: {{Towards Effective Incomplete Multi-Modal Medical Image
  Segmentation}} with {{Imbalanced Missing Rates}}.
\newblock In \emph{Proceedings of the 32nd {{ACM International Conference}} on
  {{Multimedia}}}, 456--465.

\bibitem[{Tan et~al.(2024)Tan, Zhuang, Chen, Li, Jia, Wang, and
  Li}]{tan2024epmf}
Tan, M.; Zhuang, Z.; Chen, S.; Li, R.; Jia, K.; Wang, Q.; and Li, Y. 2024.
\newblock {{EPMF}}: {{Efficient}} Perception-Aware Multi-Sensor Fusion for
  {{3D}} Semantic Segmentation.
\newblock \emph{IEEE Transactions on Pattern Analysis and Machine
  Intelligence}, 46(12): 8258--8273.

\bibitem[{Wan et~al.(2024)Wan, Zhang, Wang, Yong, Stepputtis, Sycara, and
  Xie}]{wan2024sigma}
Wan, Z.; Zhang, P.; Wang, Y.; Yong, S.; Stepputtis, S.; Sycara, K.; and Xie, Y.
  2024.
\newblock Sigma: {{Siamese}} Mamba Network for Multi-Modal Semantic
  Segmentation.

\bibitem[{Wang et~al.(2023)Wang, Ma, Zhang, Zhang, Avery, Hull, and
  Carneiro}]{wang2023learnable}
Wang, H.; Ma, C.; Zhang, J.; Zhang, Y.; Avery, J.; Hull, L.; and Carneiro, G.
  2023.
\newblock Learnable {{Cross-modal Knowledge Distillation}} for~{{Multi-modal
  Learning}} with~{{Missing Modality}}.
\newblock In \emph{Medical {{Image Computing}} and {{Computer Assisted
  Intervention}} -- {{MICCAI}} 2023: 26th {{International Conference}},
  {{Vancouver}}, {{BC}}, {{Canada}}, {{October}} 8--12, 2023, {{Proceedings}},
  {{Part IV}}}, 216--226. Berlin, Heidelberg: Springer-Verlag.
\newblock ISBN 978-3-031-43900-1.

\bibitem[{Wang et~al.(2022)Wang, Xie, Li, Fan, Song, Liang, Lu, Luo, and
  Shao}]{wang2022pvt}
Wang, W.; Xie, E.; Li, X.; Fan, D.-P.; Song, K.; Liang, D.; Lu, T.; Luo, P.;
  and Shao, L. 2022.
\newblock {{PVT}} v2: {{Improved}} Baselines with Pyramid Vision Transformer.
\newblock \emph{Computational Visual Media}, 8(3): 415--424.

\bibitem[{Wei et~al.(2025)Wei, Zhou, Lu, Yuan, and Su}]{wei2025hdbformer}
Wei, S.; Zhou, Z.; Lu, Z.; Yuan, Z.; and Su, B. 2025.
\newblock {{HDBFormer}}: {{Efficient RGB-d}} Semantic Segmentation with a
  Heterogeneous Dual-Branch Framework.
\newblock \emph{IEEE Signal Processing Letters}, 32: 91--95.

\bibitem[{Xia et~al.(2023)Xia, Zhao, Zheng, Wu, Sun, and Tang}]{xia2023cmda}
Xia, R.; Zhao, C.; Zheng, M.; Wu, Z.; Sun, Q.; and Tang, Y. 2023.
\newblock {{CMDA}}: {{Cross-modality}} Domain Adaptation for Nighttime Semantic
  Segmentation.
\newblock In \emph{2023 {{IEEE}}/{{CVF}} International Conference on Computer
  Vision ({{ICCV}})}, 21515--21524.

\bibitem[{Xie et~al.(2021)Xie, Wang, Yu, Anandkumar, Alvarez, and
  Luo}]{xie2021segformer}
Xie, E.; Wang, W.; Yu, Z.; Anandkumar, A.; Alvarez, J.~M.; and Luo, P. 2021.
\newblock {{SegFormer}}: Simple and Efficient Design for Semantic Segmentation
  with Transformers.
\newblock In \emph{Proceedings of the 35th {{International Conference}} on
  {{Neural Information Processing Systems (NeurIPS)}}}, 12077--12090.

\bibitem[{Yang et~al.(2025)Yang, Guo, Ni, Liu, Li, and Hu}]{yang2025asymmetric}
Yang, B.; Guo, Y.; Ni, R.; Liu, Y.; Li, G.; and Hu, C. 2025.
\newblock Asymmetric Multimodal Guidance Fusion Network for Realtime Visible
  and Thermal Semantic Segmentation.
\newblock \emph{Engineering Applications of Artificial Intelligence}, 142:
  109881.

\bibitem[{Zhang et~al.(2023{\natexlab{a}})Zhang, Liu, Yang, Hu, Liu, and
  Stiefelhagen}]{zhang2023cmx}
Zhang, J.; Liu, H.; Yang, K.; Hu, X.; Liu, R.; and Stiefelhagen, R.
  2023{\natexlab{a}}.
\newblock {{CMX}}: {{Cross-Modal Fusion}} for {{RGB-X Semantic Segmentation
  With Transformers}}.
\newblock \emph{IEEE Transactions on Intelligent Transportation Systems},
  24(12): 14679--14694.

\bibitem[{Zhang et~al.(2023{\natexlab{b}})Zhang, Liu, Shi, Yang, Rei{\ss},
  Peng, Fu, Wang, and Stiefelhagen}]{zhang2023delivering}
Zhang, J.; Liu, R.; Shi, H.; Yang, K.; Rei{\ss}, S.; Peng, K.; Fu, H.; Wang,
  K.; and Stiefelhagen, R. 2023{\natexlab{b}}.
\newblock Delivering Arbitrary-Modal Semantic Segmentation.
\newblock In \emph{2023 {{IEEE}}/{{CVF}} Conference on Computer Vision and
  Pattern Recognition ({{CVPR}})}, 1136--1147.

\bibitem[{Zhang et~al.(2022)Zhang, He, Yang, Li, Wei, Huang, Zhang, He, and
  Zheng}]{zhang2022mmformer}
Zhang, Y.; He, N.; Yang, J.; Li, Y.; Wei, D.; Huang, Y.; Zhang, Y.; He, Z.; and
  Zheng, Y. 2022.
\newblock {{mmFormer}}: {{Multimodal Medical Transformer}} for~{{Incomplete
  Multimodal Learning}} of~{{Brain Tumor Segmentation}}.
\newblock In \emph{2022 Medical {{Image Computing}} and {{Computer Assisted
  Intervention}} (MICCAI)}, 107--117.

\bibitem[{Zhang et~al.(2025{\natexlab{a}})Zhang, Li, Jiao, Ai, Yan, Zeng,
  Zhang, and Li}]{zhang2025cmffn}
Zhang, Y.; Li, N.; Jiao, J.; Ai, J.; Yan, Z.; Zeng, Y.; Zhang, T.; and Li, Q.
  2025{\natexlab{a}}.
\newblock {{CMFFN}}: {{An}} Efficient Cross-Modal Feature Fusion Network for
  Semantic Segmentation.
\newblock \emph{Robotics and Autonomous Systems}, 186: 104900.

\bibitem[{Zhang et~al.(2025{\natexlab{b}})Zhang, Wang, Zhu, and
  Tang}]{zhang2025tagfusion}
Zhang, Z.; Wang, W.; Zhu, L.; and Tang, Z. 2025{\natexlab{b}}.
\newblock {{TAG-fusion}}: {{Two-stage}} Attention Guided Multi-Modal Fusion
  Network for Semantic Segmentation.
\newblock \emph{Digital Signal Processing}, 156(PA).

\bibitem[{Zheng et~al.(2024{\natexlab{a}})Zheng, Lyu, Jiang, Zhou, Wang, and
  Hu}]{zheng2024magic}
Zheng, X.; Lyu, Y.; Jiang, L.; Zhou, J.; Wang, L.; and Hu, X.
  2024{\natexlab{a}}.
\newblock {{MAGIC}}++: {{Efficient}} and {{Resilient Modality-Agnostic Semantic
  Segmentation}} via {{Hierarchical Modality Selection}}.
\newblock arXiv:2412.16876.

\bibitem[{Zheng, Lyu, and Wang(2025)}]{zheng2025learninga}
Zheng, X.; Lyu, Y.; and Wang, L. 2025.
\newblock Learning {{Modality-Agnostic Representation}} for~{{Semantic
  Segmentation}} from~{{Any Modalities}}.
\newblock In \emph{2024 European Conference on Computer Vision (ECCV)},
  146--165.

\bibitem[{Zheng et~al.(2024{\natexlab{b}})Zheng, Lyu, Zhou, and
  Wang}]{zheng2025centering}
Zheng, X.; Lyu, Y.; Zhou, J.; and Wang, L. 2024{\natexlab{b}}.
\newblock Centering the~{{Value}} of~{{Every Modality}}: {{Towards Efficient}}
  and~{{Resilient Modality-Agnostic Semantic Segmentation}}.
\newblock In \emph{European Conference on Computer Vision (ECCV)}, 192--212.

\bibitem[{Zheng et~al.(2025)Zheng, Xue, Chen, Yan, Jiang, Lyu, Yang, Zhang, and
  Hu}]{zheng2025learning}
Zheng, X.; Xue, H.; Chen, J.; Yan, Y.; Jiang, L.; Lyu, Y.; Yang, K.; Zhang, L.;
  and Hu, X. 2025.
\newblock Learning {{Robust Anymodal Segmentor}} with {{Unimodal}} and
  {{Cross-modal Distillation}}.
\newblock arXiv:2411.17141.

\end{thebibliography}
